\documentclass{article}

\usepackage{arxiv}
\usepackage[utf8]{inputenc} 
\usepackage[T1]{fontenc}    
\usepackage{hyperref}       
\usepackage{url}            
\usepackage{booktabs}       
\usepackage{amsfonts}       
\usepackage{nicefrac}       
\usepackage{microtype}      
\usepackage{lipsum}
\usepackage{graphicx}
\graphicspath{ {./images/} }

\usepackage{natbib}
\usepackage{float}
\usepackage{booktabs}
\usepackage{multirow}
\usepackage{bm}
\usepackage[ruled,vlined]{algorithm2e}
\usepackage{pifont}
\newcommand{\cmark}{\text{\ding{51}}}

\usepackage{amsfonts}

\newcommand{\argmax}{\arg\!\max}
\DeclareUnicodeCharacter{0301}{\'{e}}
\usepackage{mathtools, nccmath}

\title{Semi-Supervised Hierarchical Drug Embedding in Hyperbolic Space}

\author{
 Ke Yu \\
  School of Coumputing and Information\\
  University of Pittsburgh\\
  Pittsburgh, PA 15206 \\
  \texttt{yu.ke@pitt.edu} \\
   \And
 Shyam Visweswaran \\
  Department of Biomedical Informatics\\
  University of Pittsburgh\\
  Pittsburgh, PA 15206 \\
  \texttt{shv3@pitt.edu} \\
  \And
 Kayhan Batmanghelich \\
  Department of Biomedical Informatics\\
  University of Pittsburgh\\
  Pittsburgh, PA 15206 \\
  \texttt{kayhan@pitt.edu} \\
}

\begin{document}
\maketitle
\begin{abstract}
Learning accurate drug representations is essential for tasks such as computational drug repositioning and prediction of drug side-effects. A drug hierarchy is a valuable source that encodes human knowledge of drug relations in a tree-like structure where drugs that act on the same organs, treat the same disease, or bind to the same biological target are grouped together. However, its utility in learning drug representations has not yet been explored, and currently described drug representations cannot place novel molecules in a drug hierarchy.
\\
Here, we develop a semi-supervised drug embedding that incorporates two sources of information: (1) underlying chemical grammar that is inferred from molecular structures of drugs and  drug-like molecules (unsupervised), and (2) hierarchical relations that are encoded in an expert-crafted hierarchy of approved drugs (supervised). We use the Variational Auto-Encoder (VAE) framework to encode the chemical structures of molecules and use the knowledge-based drug-drug similarity to induce the clustering of drugs in hyperbolic space. The hyperbolic space is amenable for encoding hierarchical concepts. Both quantitative and qualitative results support that the learned drug embedding can accurately reproduce the chemical structure and induce the hierarchical relations among drugs. Furthermore, our approach can infer the pharmacological properties of novel molecules by retrieving similar drugs from the embedding space. We demonstrate that the learned drug embedding can be used to find new uses for existing drugs and to discover side-effects. We show that it significantly outperforms baselines in both tasks.
\end{abstract}

\section{Introduction}

The study of drug representation provides the foundation for a variety of applications in computational pharmacology, such as computational drug repositioning and prediction of drug side-effects. Drug repositioning, the process of finding new uses for existing drugs, is one strategy to shorten the time and reduce the cost of drug development~\citep{nosengo2016new}. Computational methods of drug repositioning typically aim to identify shared mechanism of actions among drugs that imply that the drugs may also share therapeutic applications~\citep{pushpakom2019drug}. However, such methods are limited when prior knowledge of drugs may be scarce or not available; for example, drugs that are in the experimental phase or have failed clinical trials. Therefore, it is appealing to map the chemical structure of a molecule to its pharmacological behavior. Side-effects of drugs are undesirable effects that are harmful to patients and can even be fatal. Computational methods for predicting drug side-effects often integrate several drug features from heterogeneous data sources (e.g., chemical, biological and therapeutic properties)~\citep{liu2012large}. However, the utility of drug hierarchy in learning drug representation has not yet been explored. A drug hierarchy encodes a broad spectrum of known drug relations. For example, a widely used drug hierarchy, Anatomical Therapeutic Chemical Classification System (ATC), groups drugs that are similar in terms of their mechanism of action and therapeutic, pharmacological and chemical characteristics.

Representing the chemical structure of drug-like molecules has received substantial attention recently~\citep{walters2020assessing}. This approach focuses on learning representations that can be used to identify promising molecules that satisfy specified properties~\citep{yang2019analyzing, segler2018generating, gomez2018automatic}. Typically, a large set of drug-like molecules is encoded in a latent space, which is then coupled with a predictive model. However, this approach does not directly incorporate prior knowledge about existing drugs.
In another approach, knowledge about existing drugs is leveraged to predict hitherto unknown properties of drugs. Each drug as denoted as a node in graph and linkages between drugs are predicted where a linkage may indicate a new use~\citep{yu2016prediction}, a side effect~\citep{timilsina2019discovering}, or an adverse drug-drug interaction~\citep{zitnik2018modeling}. However, such an approach is limited to the drugs available in the knowledge database and learns task-specific representations that may not transfer well to additional tasks. Our method merges the two approaches described above by combining chemical structure representation learning with known knowledge of drugs to learn useful and generalizable drug representation.

\begin{figure*}[ht]
    \centering
    \includegraphics[width=1.0\textwidth]{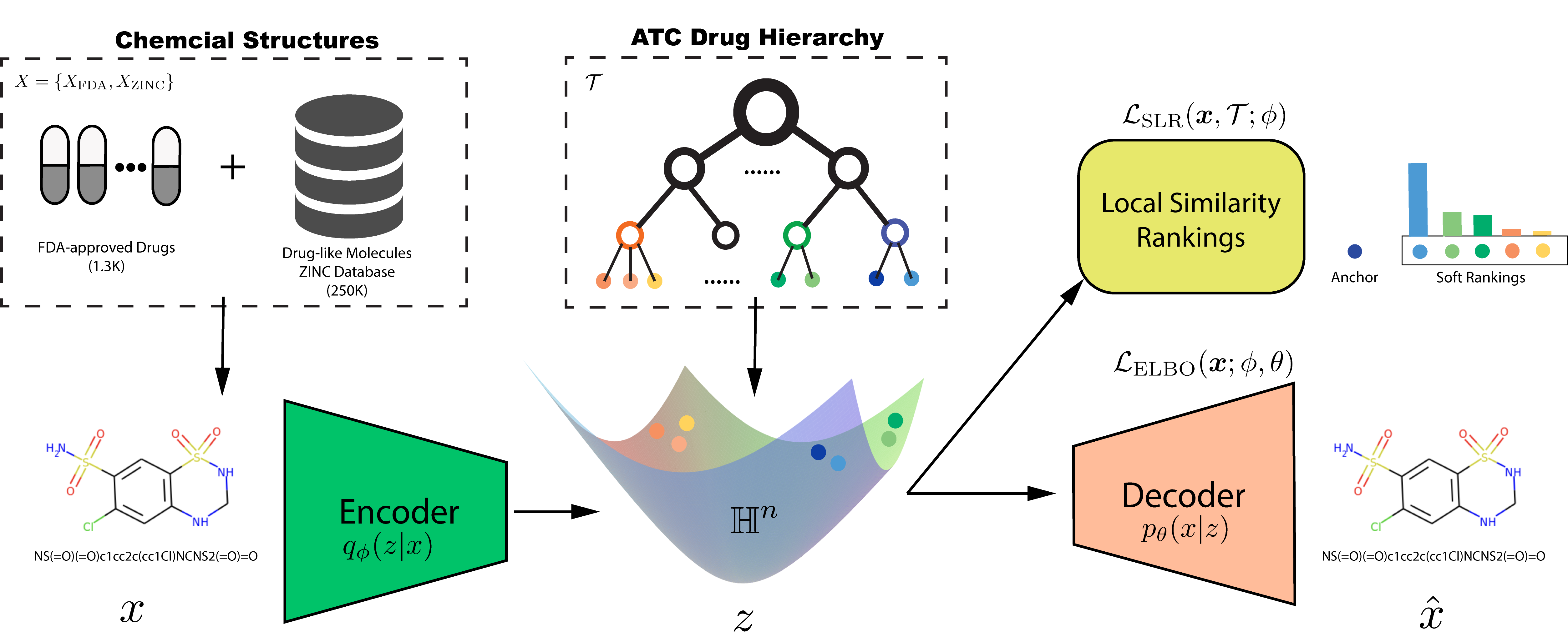}
    \caption{Schematic diagram of the proposed drug embedding method. Our semi-supervised learning approach integrates the chemical structures of a small number of FDA-approved drug molecules ($X_\mathrm{FDA}$) and a larger number of drug-like molecules ($X_\mathrm{ZINC}$) drawn from the ZINC database. We use VAE to encode molecules in hyperbolic space $\mathbb{H}^n$, and enforce the ATC drug hierarchy by preserving local similarity rankings of drugs. The symbols $\bm{x}$, $\bm{z}$, $\hat{\bm{x}}$ denote a molecule represented by its SMILES string, its embedding and its reconstruction; $q_\phi(\bm{z}|\bm{x}), p_\theta(\bm{x}|\bm{z})$ denote the encoder network and the decoder network respectively; $\mathcal{L}_\mathrm{ELBO}(\bm{x};\phi,\theta)$, $\mathcal{L}_\mathrm{SLR}(\bm{x},\mathcal{T};\phi)$ denote the objective functions for the VAE and the local similarity rankings.}
    \label{fig:1}
\end{figure*}

\noindent \textbf{Problem setup.} Here, we develop a drug embedding that integrates the chemical structures of drug-like molecules with a drug taxonomy such that the similarity between pairs of drugs is informed both by the structure and groupings in the taxonomy (Figure~\ref{fig:1}). To learn the underlying grammar of chemical structures, we leverage a data set of drugs (about 1.3K) that are approved by the Food and Drug Administration (FDA) and a larger data set of drug-like molecules (about 250K) and use the simplified molecular-input line-entry system (SMILES)~\citep{weininger1988smiles} structure representation. We obtain additional drug similarity relationships from the ATC drug taxonomy that hierarchically groups drugs by the system of action, therapeutic intent, pharmacological and chemical characteristics. We use the hyperbolic space for the embedding since it is amenable for learning continuous concept hierarchies~\citep{nickel2017poincare, mathieu2019continuous,de2018representation, monath2019gradient}.

We formulate the learning of the drug embedding as a Variational Auto-Encoder where the codes ($\bm{z}$) reside in hyperbolic space. More specifically, we adopt a variant that replaces the prior normal distribution in the VAE with the so-called wrapped normal distribution in the Lorentz model of hyperbolic space to leverage its numerical stability~\citep{nickel2018learning, nagano2019wrapped}. To integrate the hierarchical relationships from the drug taxonomy, we use a loss function that enforces the pairwise hyperbolic distance between drugs to be consistent with pairwise shortest path lengths in the ATC tree. 


\section{Background}
\label{sec:background}
In this section, we provide a brief introduction to the hyperbolic space and the Lorentz model of hyperbolic geometry.
Hyperbolic space is a geometric space that is characterized by constant negative curvature and in which Euclid’s parallel postulate does not hold. Intuitively, hyperbolic space can be viewed as a continuous version of tree for its volume and surface area grow exponentially with radius. Compared to Euclidean space, hyperbolic space better captures the hierarchical characteristic of trees. In this paper, we employ a specific model of hyperbolic space, namely, the Lorentz (Minkowski/Hyperboloid) model. This model is computationally appealing because it has a simple closed-form distance function and has analytic forms for Exponential map, Logarithm map, and parallel transport. 

\textbf{Lorentz model.} The Lorentz model $\mathbb{H}^n$ of $n$-dimensional hyperbolic space is defined as follows:
\begin{equation}
    \mathbb{H}^n = \{\bm{z} \in \mathbb{R}^{n+1}: \langle \bm{z} \,,\bm{z} \rangle_{\mathcal{L}} = -1, z_0 > 0 \}
\end{equation}
where $\bm{z}, \bm{z}' \in \mathbb{R}^{n+1}$, and $
    \langle \bm{z} \,,\bm{z}' \rangle_{\mathcal{L}} = -z_0z'_0 + \displaystyle \textstyle \sum_{i=1}^{n}z_iz'_i$ is the so-called the \emph{Lorentzian inner product}, which is also the metric tensor of the hyperbolic space. The origin in the hyperbolic space is $\bm{\mu}_0=[1,0,0,...,0] \in \mathbb{H}^n$ and the Lorentzian inner product of any point $\bm{z} \in \mathbb{H}^n$ with itself is $-1$. 
 
\textbf{Geodesic.} A geodesic between two points $\bm{z}, \bm{z}' \in \mathbb{H}^n$ is a curve representing the shortest path between these two points. Fortunately, the geodesic distance, $d_{\ell}(\bm{z}, \bm{z}')$, has a closed-form in $\mathbb{H}^n$:
\begin{equation}
\label{eq:lor_dist}
    d_{\ell}(\bm{z}, \bm{z}') = \mathrm{cosh}^{-1} (-\langle \bm{z} \,,\bm{z}' \rangle_{\mathcal{L}}).
\end{equation}
where $\mathrm{cosh}^{-1}$ denotes inverse of hyperbolic cosine function. 

\textbf{Tangent space.} The tangent space $T_{\bm{\mu}}\mathbb{H}$ can be viewed as a linear approximation of the manifold around the point $\bm{\mu} \in \mathbb{H}^n$. Formally, $T_{\bm{\mu}}\mathbb{H}$ is a set of vectors that take paths in any direction running through the point $\bm{\mu}$ and are hyperbolic orthogonal to $\bm{\mu}$:
\begin{equation}
    T_{\bm{\mu}}\mathbb{H} := \{ \bm{u}: \langle \bm{u}, \bm{\mu} \rangle_{\mathcal{L}} = 0 \}.
\end{equation}
When $\bm{\mu}=\bm{\mu}_0$, the origin, $T_{\bm{\mu}_0}\mathbb{H}$ consists of vectors $\bm{u}$ with $\|\bm{u}\|_2 = \|\bm{u}\|_{\mathcal{L}}$. This implies that the $\ell_2$ norm of any vector of the tangent space of the origin is the same as its metric norm. 

\textbf{Exponential and Logarithmic maps.} The \textit{exponential map}, 
$\mathrm{exp}_{\bm{\mu}}:T_{\bm{\mu}}\mathbb{H} \rightarrow \mathbb{H}^n$, projects vectors from the tangent space to the hyperbolic space. In general, the \textit{exponential map} is only defined locally, that is, it maps only a small neighborhood of the origin in $T_{\bm{\mu}}\mathbb{H}$, to a neighborhood of $\bm{\mu}$ in the manifold $\mathbb{H}^n$. For any vector $\bm{u} \in T_{\bm{\mu}}\mathbb{H}$, we can map $\bm{u}$ to a point $\bm{z}$ so that the hyperbolic distance between $\bm{\mu}$ and $\bm{z}$ matches with $\|\bm{u}\|_\mathcal{L}$. The \textit{exponential map} of the Lorentz model is defined as:
\begin{equation}
\label{eq:expmap}
    \bm{z} = \mathrm{exp}_{\bm{\mu}}(\bm{u}) = \mathrm{cosh}(\|\bm{u}\|_\mathcal{L})\bm{\mu} + \mathrm{sinh}(\|\bm{u}\|_\mathcal{L}) \frac{\bm{u}}{\|\bm{u}\|_\mathcal{L}}
\end{equation}
The inverse exponential map, also known as \textit{logarithmic map}, is defined as:
\begin{equation}
\label{eq:invexpmap}
    \bm{u} = \mathrm{exp}_{\bm{\mu}}^{-1}(\bm{z}) = \frac{\mathrm{cosh}^{-1}(\gamma)}{\sqrt{\gamma^2-1}}(\bm{z}-\gamma\bm{\mu})
\end{equation}
where $\gamma = - \langle \bm{\mu}, \bm{z} \rangle_\mathcal{L}$.

\section{Method}
\label{sec:method}
\subsection{Learning chemical grammar using VAE}
We use a variational autoencoder (VAE) to encode the chemical structure of drug-like molecules~\citep{gomez2018automatic}. More specifically, we model a molecule as a random variable generated by encoding a SMILES string into a code ($\bm{z}$), which is then decoded back to a reconstruction of the input by passing through a decoder $p_{\theta}(\bm{x}| \bm{z})$. Finding the optimal $\theta$ using maximum likelihood requires computing the so-called evidence function, $\log p_{\theta}(\bm{x})$, which is difficult to compute since it entails integrating over $\bm{z}$.

Instead of directly maximizing likelihood, variational Bayes maximizes the variational lower bound, also known as ELBO~\citep{jordan1999introduction}. The ELBO is given by 
\begin{equation}
  \log p_{\theta}(\bm{x}) \geq \mathbb{E}_{\bm{z} \sim q(\bm{z}| \bm{x}) } \left[ \log p_{\theta}( \bm{x} | \bm{z} ) \right] - D_\mathrm{KL} \left( q( \bm{z} | \bm{x}) || p(\bm{z}) \right)
\end{equation} 
where the first term after the inequality is the reconstruction term and the second term is the regularization term, and $\mathbb{E}$ and $D_\mathrm{KL}$ denote the expectation and Kullback–Leibler (KL) divergence respectively. The global optimal $q( \bm{z} | \bm{x})$ is achieved when $q( \bm{z} | \bm{x}) = p(\bm{z} | \bm{x})$, the variational distribution approximates the posterior distribution. In order to control the relative effect of KL divergence~\citep{he2019lagging} we adopt $\beta$-VAE~\citep{higgins2017beta}, a more general form of VAE that applies a scaling hyperparameter $\beta$ to the $D_\mathrm{KL}$ term in the ELBO.
We employ the recurrent neural network (RNN)~\citep{cho2014learning} architecture for both the encoder and the decoder networks, in order to perform sequence-to-sequence learning on SMILES strings.

In the classic VAE~\citep{kingma2013auto}, the prior $p(\bm{z})$ is the standard normal distribution, the encoder $q_\phi(\bm{z}| \bm{x})$ is modeled by a Gaussian distribution $\mathcal{N}(\bm{z}|\bm{\mu}_\phi, \Sigma_\phi)$, and the first term in the ELBO is estimated using a Monte Carlo estimator:
\begin{equation}
    \mathbb{E}_{\bm{z} \sim q_\phi(\bm{z}| \bm{x}_i) } \left[\mathrm{log}p_\theta(\bm{x}_i| \bm{z}) \right] \approx \frac{1}{L} \sum_{l=1}^{L} \mathrm{log}p_\theta(\bm{x}_i | g_\phi(\bm{\epsilon}^{(l)}, \bm{x}_i))
\end{equation}
where $g_\phi(\bm{\epsilon}^{(l)}, \bm{x}_i) = \bm{\mu}_{\phi}^{(i)} + \bm{\sigma}_{\phi}^{(i)} \odot \bm{\epsilon}^{(l)}$ and $\bm{\epsilon}^{(l)} \sim \mathcal{N}(0, \bm{\mathrm{I}})$ is the reparameterization trick, and $L$ is the number of samples per data point. To extend VAE from a flat Euclidean space to a curved manifold, the Gaussian distribution should be extended to the hyperbolic manifold. Also, the reparameterization trick should be modified as algebraic addition of coordinates of two points on a manifold does not necessarily reside on the manifold. 

\textbf{Wrapped normal.} We adopt the so-called wrapped normal distribution proposed by Nagano et al., 2019~\citep{nagano2019wrapped}, which we denote by $\mathcal{N}_{\mathbb{H}}^{\mathrm{W}}(\bm{z}|\bm{\mu}, \Sigma)$, where $\bm{z} \in \mathbb{H}^n$ and $\bm{\mu}$ is the hyperbolic mean. The sampling strategy can be summarized in three steps. First, we define a Gaussian random variable, $\bm{u} \sim \mathcal{N}(\bm{0},\Sigma)$, on the tangent space at the origin of the hyperbolic space, $\bm{u} \in T_{\bm{\mu_0}}\mathbb{H}$. Then, we \textit{parallel transport} the random vector to another tangent space at a desired location $\bm{\mu}$. The \textit{parallel transport} translates a vector from $T_{\bm{\mu_0}}\mathbb{H}$ to $T_{\bm{\mu}}\mathbb{H}$ along the geodesic between $\bm{\mu}_0$ and $\bm{\mu}$ without changing its metric tensor. The formula for parallel transport in the Lorentz model is given by~\citep{helgason2001differential}:
\begin{equation}
\label{eq:pt2}
    \bm{u}_\mathrm{pt} = \mathrm{PT}_{\bm{\mu}_0 \rightarrow \bm{\mu}} (\bm{u}) = \bm{u}+ \frac{\langle \bm{\mu} - \alpha \bm{\mu}_0, \bm{u} \rangle _\mathcal{L}}{\alpha + 1}(\bm{\mu}_0 + \bm{\mu})
\end{equation}
where $\alpha = - \langle \bm{\mu}_0, \bm{\mu} \rangle _\mathcal{L}$ and $\bm{u}_\mathrm{pt} \in T_{\bm{\mu}}\mathbb{H}$. Finally, we map the transported vector into hyperbolic space via the \textit{exponential map}, $\bm{z} = \mathrm{exp}_{\bm{\mu}}(\bm{u}_\mathrm{pt})$. Importantly, this sampling scheme is sequentially norm-preserving, i.e., $\|\bm{u}\|_2=\|\bm{u}\|_\mathcal{L}=\|\bm{u}_\mathrm{pt}\|_\mathcal{L}= \|\bm{z}\|_\mathcal{L}$. 

\textbf{Reparameterization trick.} 
The composition of these two operations, $\mathrm{exp}_{\bm{\mu}}(\mathrm{PT}_{\bm{\mu}_0 \rightarrow \bm{\mu}} (\bm{u}))$, can be viewed as the reparameterization trick in the hyperbolic VAE. The inside operation shifts the tangent space from $\bm{\mu}_0$ to $\bm{\mu}$ analogous to the addition operation of the classic reparameterization trick. The $\mathrm{exp}_{\bm{\mu}}$ project the shifted vector to the manifold. Therefore, we sample $\bm{z}_i^{(l)} \sim q_\phi(\boldsymbol{z}|\boldsymbol{x_i})$ using:
\begin{equation}
\label{eq:hrep}
    \bm{z}_i^{(l)} = g_\phi(\bm{u}^{(l)}, \bm{\mu}_i) = \mathrm{exp}_{\bm{\mu}_i}(\mathrm{PT}_{\bm{\mu}_0 \rightarrow \bm{\mu}_i} (\bm{u}^{(l)}))
\end{equation}
where $\bm{u}^{(l)} \sim \mathcal{N}(\bm{0},\Sigma)$, and $l$ denotes the index of sample. Note that, in the Lorentz model, both the \textit{parallel transport} and the \textit{exponential map} have analytical forms, i.e., Eq.~(\ref{eq:pt2}),~(\ref{eq:expmap}), and can be differentiated with respect to the hyperbolic mean $\bm{\mu}$ of the wrapped normal distribution $\mathcal{N}_{\mathbb{H}}^{\mathrm{W}}(\bm{z}|\bm{\mu}, \Sigma)$. 

\textbf{KL Divergence.}
To compute the KL divergence, we need to evaluate the probability density of the wrapped normal. The wrapped normal distribution can be viewed as change of variable from a normal distribution via the Eq.~(\ref{eq:hrep}). Applying the change of variable, we arrive at:
\begin{equation}
\label{eq:KL1}
       \mathrm{log}\ q_\phi(\bm{z}_i^{(l)}|\bm{x}_i)
        = \mathrm{log}\ \mathcal{N}\Big(g_\phi^{-1}(\bm{z}_i^{(l)}, \bm{\mu}_i); \bm{0}, \Sigma \Big)
        - \mathrm{log}\det \bigg(\frac{\partial g_\phi(\bm{u}^{(l)}, \bm{\mu}_i)}{\partial \bm{u}^{(l)}} \bigg)
\end{equation}
The inverse operation $g_\phi^{-1}(\bm{z}_i^{(l)}, \bm{\mu}_i)$ simply maps $\bm{z}_i^{(l)}$ back to $\bm{u}^{(l)}$ by applying the \textit{logarithmic map}, Eq.~(\ref{eq:invexpmap}) and the \textit{inverse parallel transport}, $\mathrm{PT}_{\bm{\mu}_i \rightarrow \bm{\mu}_0} (\bm{u}_\mathrm{pt}^{(l)})$. We compute the second log-determinant term following the derivation in~\citep{nagano2019wrapped}.

\subsection{Integrating hierarchical knowledge}
\label{sec:slr}
\textbf{Soft local ranking.} The hyperbolic VAE learns an embedding for codes that are amenable to hierarchical representation. However, it only models $\bm{x}$ (the SMILES string of the drug), and it does not enforce our prior knowledge about drug taxonomy, which defines similarity or dissimilarity between drugs at various levels. In this section, we incorporate the ATC hierarchy into our model. Note that the terminal nodes of the ATC hierarchy are drugs that have SMILES string representations, while the internal nodes of the ATC hierarchy are drug classes, e.g., beta blocking agents. Inspired by concept embedding in hyperbolic space~\citep{nickel2018learning}, we incorporate the ATC hierarchy in our model by using pairwise similarity between drugs.
Let $t_{i,j}$ denote the path-length between two drugs, $\bm{x}_{i}$ and  $\bm{x}_{j}$ in $\mathcal{T}$, and let $\mathcal{D}(i,j)=\{k: t_{i,j}<t_{i,k}\} \cup \{j\}$ denote the set of drugs with path-lengths equal to or greater than $t_{i,j}$. We define the soft local ranking with respect to the anchor drug $\bm{x}_i$ as:
\begin{equation}
\label{eq:slr}
    p(\bm{x}_i, \bm{x}_j;\phi) = \frac{\mathrm{exp}(-d_{\ell}(\bm{\mu}_i, \bm{\mu}_j))}{\sum_{k \in \mathcal{D}(i,j)} \mathrm{exp}(-d_{\ell}(\bm{\mu}_i, \bm{\mu}_k))}
\end{equation}
where $\bm{\mu}_i$ is the hyperbolic mean of $q_\phi(\bm{z}|\bm{x}_i) = \mathcal{N}_{\mathbb{H}}^{\mathrm{W}}(\bm{z}|\bm{\mu}_i, \Sigma_i)$ and $d_{\ell}(\bm{\mu}_i, \bm{\mu}_j)$ is the hyperbolic distance between $\bm{\mu}_i$ and $\bm{\mu}_j$. The likelihood function of the soft local rankings per $\bm{x}_i \in X_\mathrm{FDA}$ is given by:
\begin{equation}
\label{eq:slr_est}
    \mathcal{L}_{\mathrm{SLR}}(\bm{x}_i, \mathcal{T}; \phi) = \sum_j \mathrm{log}\ 
    p(\bm{x}_i, \bm{x}_j;\phi)
\end{equation}
where $\bm{x}_j \in \{X_\mathrm{FDA} - \bm{x}_i\}$.

Note that the global hierarchy of $\mathcal{T}$ is decomposed into local rankings denoted by $\mathcal{D}(i,j)=\{k: t_{i,j}<t_{i,k}\} \cup \{j\}$. To train our model, we need to effectively sample $\mathcal{D}(i,j) \sim \mathcal{T}$, and the best sampling strategy supported by the results of our experiments is as follows. For each anchor drug $\bm{x}_i$, we uniformly sample a positive example $\bm{x}_j$, such that the lowest common ancestor of $\bm{x}_i$, $\bm{x}_j$ has equal chance of being an internal node at any level, i.e., level 1, 2, 3, or 4, in the ATC tree. We then randomly sample $k$ negative examples $\bm{x}_k$ from other leaf nodes that have greater path lengths than $t_{i,j}$. 

\subsection{Optimization}
\textbf{Formulation.} We employ a semi-supervised learning approach that combines a small number of drugs $X_\mathrm{FDA}$ with a larger number of drug-like molecules $X_\mathrm{ZINC}$. The supervised learning task is to maximize the likelihood of the soft local rankings with respect to the ATC hierarchy $\mathcal{T}$. The unsupervised learning task is to maximize the ELBO of the marginal likelihood of the chemical structures of drugs and drug-like molecules $X=\{X_\mathrm{ZINC}, X_\mathrm{FDA}\}$. We then formulate the drug embedding problem as:
\begin{equation}
\label{eq:obj}
     \argmax_{\phi, \theta} \Big(\mathcal{L}_{\beta-\mathrm{ELBO}}(\bm{x}; \phi, \theta)
      + c\cdot\mathcal{L}_{\mathrm{SLR}}(\bm{x}, \mathcal{T}; \phi) \Big)
\end{equation}
where $c=1$ when $\bm{x} \in X_\mathrm{FDA}$, $c=0$ when $\bm{x} \in X_\mathrm{ZINC}$, and $|X_\mathrm{ZINC}| \gg |X_\mathrm{FDA}|$. The first term in the objective function captures the underlying chemical grammar of molecules, and the second term enforces the relative positions of the drugs in the latent space to correspond to their relative positions in the ATC hierarchy.

\textbf{Training.} In practice, the learning procedure for the parameters $\phi$, $\theta$ is summarized as:
\begin{equation}
\label{eq:lossfunc}
   \argmax_{\phi,\theta}  \frac{1}{|X|} \displaystyle\sum_{\bm{x}_i \in X}
   \Big( \mathrm{log}\ p_\theta(\bm{x}_i|\bm{z}_i) - \beta \cdot \Tilde{D}_\mathrm{KL}\left(q_\phi(\boldsymbol{z}_i|\boldsymbol{x}_i)\|p(\boldsymbol{z}_i) \right)\Big) 
    + \gamma\cdot \frac{1}{|X_\mathrm{FDA}|} \displaystyle\sum_{\bm{x}_i \in X_\mathrm{FDA}}\Tilde{\mathcal{L}}_{\mathrm{SLR}}(\bm{x}_i, \mathcal{T}; \phi)
\end{equation}
where $\bm{z}_i$ is a single sample, and $\beta$ and $\gamma$ are scaling hyperparameters governing the relative weights of KL divergence and soft local ranking loss during training. Parameters are estimated using stochastic gradient descent, and gradients are straightforward to compute using the hyperbolic reparameterization trick Eq.~(\ref{eq:hrep}). For details of model architectures, training settings and other implementation details, please refer to the Supplementary Material.

\section{RELATED WORK}
\label{sec:relatedworks}
Substantial research has been done in the past few years in applying machine learning to drug discovery and related tasks~\citep{butler2018machine, vamathevan2019applications, ekins2019exploiting}. Examples of applications include extracting information from the chemical structure to search molecules with desirable properties~\citep{gomez2018automatic}, predicting drug toxicity~\citep{unterthiner2015toxicity}, and identifying novel targets for drugs~\citep{jeon2014systematic}. Machine learning has also been applied to predict side-effects of using multiple drugs~\citep{zitnik2018modeling} and drug repositioning~\citep{yu2016prediction}. However, many of these methods represent a drug as a node in a graph, and ignore the rich information in the chemical structure of the drug. Moreover, such approaches are unable to analyzing a new drug that is not already in the data set. Our method can be viewed as knowledge representation learning that integrates information from (1) a large corpus of drug-like molecules that is used in drug discovery, and (2) an expert-curated drug taxonomy that embeds rich information about known drugs. To the best of our knowledge, our method is the first approach that allows localizing novel molecules in the context of the already clinically approved drugs.

Molecular featurization methods can be divided into two groups: (1) methods that extract expert-crafted features from molecular structures such as extended-connectivity fingerprints~\citep{rogers2010extended}, coulomb matrix~\citep{rupp2012fast}, and (2) recent deep learning based methods. The deep learning-based methods can be further categorized into two groups. The first group consists of methods that encode the molecular formula as a string of characters and use a variant of RNN to extract features~\citep{gomez2018automatic, kusner2017grammar, gupta2018generative}, and the second group contains methods that represent as undirected graphs where nodes are atoms and edges are bonds~\citep{simonovsky2018graphvae, liu2018constrained}. Each group has advantages and disadvantages~\citep{alperstein2019all}. Our method belongs to the first group of deep learning-based methods. However, our framework is quite general in that the encoder-decoder can be replaced with a graph-neural encoder-decoder if needed.

Embedding hierarchical concepts in a latent space has been an active area of research~\citep{goyal2017nonparametric, monath2019gradient}. Hyperbolic space is an appealing choice for embedding a hierarchy because it can represent tree-like structures with arbitrarily low distortion~\citep{de2018representation}. There are several equivalent geometric models~\citep{helgason2001differential} of hyperbolic space. Many applications of hyperbolic space to machine learning~\citep{nickel2017poincare, mathieu2019continuous, monath2019gradient} have adopted the Poincar\'e ball model. However, as claimed in~\cite{nickel2018learning}, the Lorentz model allows for a more efficient closed-form computation of geodesics and avoids numerical instabilities that arise from the Poincare distance. A more recent study~\citep{nagano2019wrapped} introduced the \textit{wrapped normal distribution} in the Lorentz model. To the best of our knowledge, our work is the first hyperbolic VAE framework which can induce hierarchical structure from pairwise similarity measurements in a latent space.

\section{Experiments}
\label{sec:experiments}
In this section, we first describe the datasets used in our experiments. Then, we perform three sets of experiments to evaluate different components of our model: (1) effect of the ATC information in preserving hierarchical relations among drugs, and (2) importance of hyperbolic space as the coding space. 
Finally, we study the efficacy of hyperbolic embeddings for (1) discovering side-effects of drugs and (2) drug repositioning.

\subsection{Datasets}
\textbf{Chemical structures.} We obtained SMILES strings of 1,365 FDA-approved drugs that were curated by~\citep{gomez2018automatic}. We obtained SMILES strings of 250,000 drug-like molecules that were extracted at random by~\citep{gomez2018automatic} from the ZINC database that contains a curated collection of >200M commercially available chemicals. We combine the 1,365 drug and the 250,000 drug-like molecules to create a single data set of molecular structures that we use in our experiments.

\textbf{ATC.} The ATC taxonomy was created by the World Health Organization (WHO)~\citep{world2014collaborating} that leverages the location of action, therapeutic, pharmacological and chemical properties of drugs to group them hierarchically. Traversing from the top to the bottom of the hierarchy, the ATC groups drugs according to the anatomical organ on which they act (level 1), therapeutic intent (level 2), pharmacological properties (level 3) and chemical characteristics (level 4). A drug that has several uses appears in several places in the ATC hierarchy. We obtained the ATC hierarchy from the UMLS Metathesaurus (version 2019AB) and mapped the FDA-approved drugs to the terminal nodes in the ATC tree that represents the active chemical substance (level 5). Of the 1,365 drugs, 1,055 were mapped to 1,355 terminal nodes at level 5 in the ATC tree.

\textbf{SIDER.} The Side Effect Resource (SIDER) database~\citep{kuhn2016sider} contains 5,868 distinct side effects and 1,427 drugs for which one or more side effects have been documented. We obtained the SIDER data set in DeepChem~\citep{wu2018moleculenet}, which has grouped side effects into 27 classes based on the anatomical organ that is affected by the side effect.

\textbf{RepoDB.} RepoDB~\citep{brown2017standard} is a benchmark data set that contains information on drug repositioning. It contains a curated set of drug repositioning successes and failures where each success or failure is a drug-indication pair where indication refers to a specific condition that the drug is used to treat. After mapping to FDA-approved drugs, we obtained 4,738 successful and 2,576 failed drug-indication pairs. 

\subsection{Evaluating drug embeddings}
\label{sec:lde_quality}
\label{sec:exp_metric}
We assess the quality of hyperbolic embeddings in their ability to accurately capture the chemical structure as well as preserve relationships entailed by the ATC hierarchy. To learn embeddings, we randomly split the chemical structures data set into training, validation and test sets in the proportions 90\%:5\%:5\%. The validation set is used to determine the best-fit model.

\textbf{Metrics.} We evaluate the embeddings in their ability to recapitulate the ATC hierarchy by applying agglomerative hierarchical clustering to the embeddings. We compare the embedding-induced hierarchy to the ATC hierarchy using dendrogram purity~\citep{heller2005bayesian}. The dendrogram purity (DP) of a hierarchy $\Tilde{\mathcal{T}}$ that is obtained from a set of drug embeddings $\{\bm{\mu}_i\}$ is computed as:
\begin{equation}
\label{eq:dp}
    \mathrm{DP}(\Tilde{\mathcal{T}}) = \frac{1}{|\mathcal{W}\star|} \sum_{\bm{\mu}_i,\bm{\mu}_j \in  \mathcal{W}^{\star}}\mathrm{pur}\big(\mathrm{lvs}(\mathrm{LCA}(\bm{\mu}_i,\bm{\mu}_j)), \mathcal{C}^{\star}(\bm{\mu}_i)\big)
\end{equation}
where $\mathcal{C}^{\star}(\bm{\mu}_i)$ is the (ground-truth) cluster that the drug $\bm{x}_i$ belongs to in the ATC $\mathcal{T}$, $\mathcal{W}^{\star}$ is the set of unordered pairs of drugs that belong to the same cluster, $\mathrm{LCA}(\bm{\mu}_i,\bm{\mu}_j)$ is a function that gives the lowest common ancestor of $\bm{\mu}_i$ and $\bm{\mu}_j$ in $\Tilde{\mathcal{T}}$, $\mathrm{lvs}(n)$ is the set of descendant leaves for any internal node $n$ in $\Tilde{\mathcal{T}}$, and $\mathrm{pur}(\bm{S}_1,\bm{S}_2) = |\bm{S}_1 \cap \bm{S}_2|/|\bm{S}_1|$. Intuitively, DP measures the average purity of the lowest common ancestors of pairs of drugs that belong to the same ATC cluster. Note that $\mathrm{DP}(\Tilde{\mathcal{T}})$ is a holistic measure of the complete ATC hierarchy that includes drugs in the training set. 

We also evaluate how well the embeddings are decoded to the original SMILES strings. Following~\citep{gomez2018automatic}, we evaluate the reconstruction accuracy as the proportion of successful decoding of latent representation after 100 attempts for 1,000 molecules randomly chosen from the test set. 

\textbf{Effect of knowledge source.}
We evaluate DP and reconstruction accuracy of embeddings obtained from a single source of knowledge that includes (1) chemical structures only by maximizing $\mathcal{L}_{\beta-\mathrm{ELBO}}(\bm{x}; \phi, \theta)$ using the entire $X$, and (2) ATC hierarchy only by  maximizing $\mathcal{L}_{\mathrm{SLR}}(\bm{x}, \mathcal{T}; \phi)$ using $X_\mathrm{FDA}$. We compare them to the embedding that is obtained from both chemical structures and ATC hierarchy.

\begin{figure}[h]
    \centering
    \includegraphics[width=0.8\textwidth]{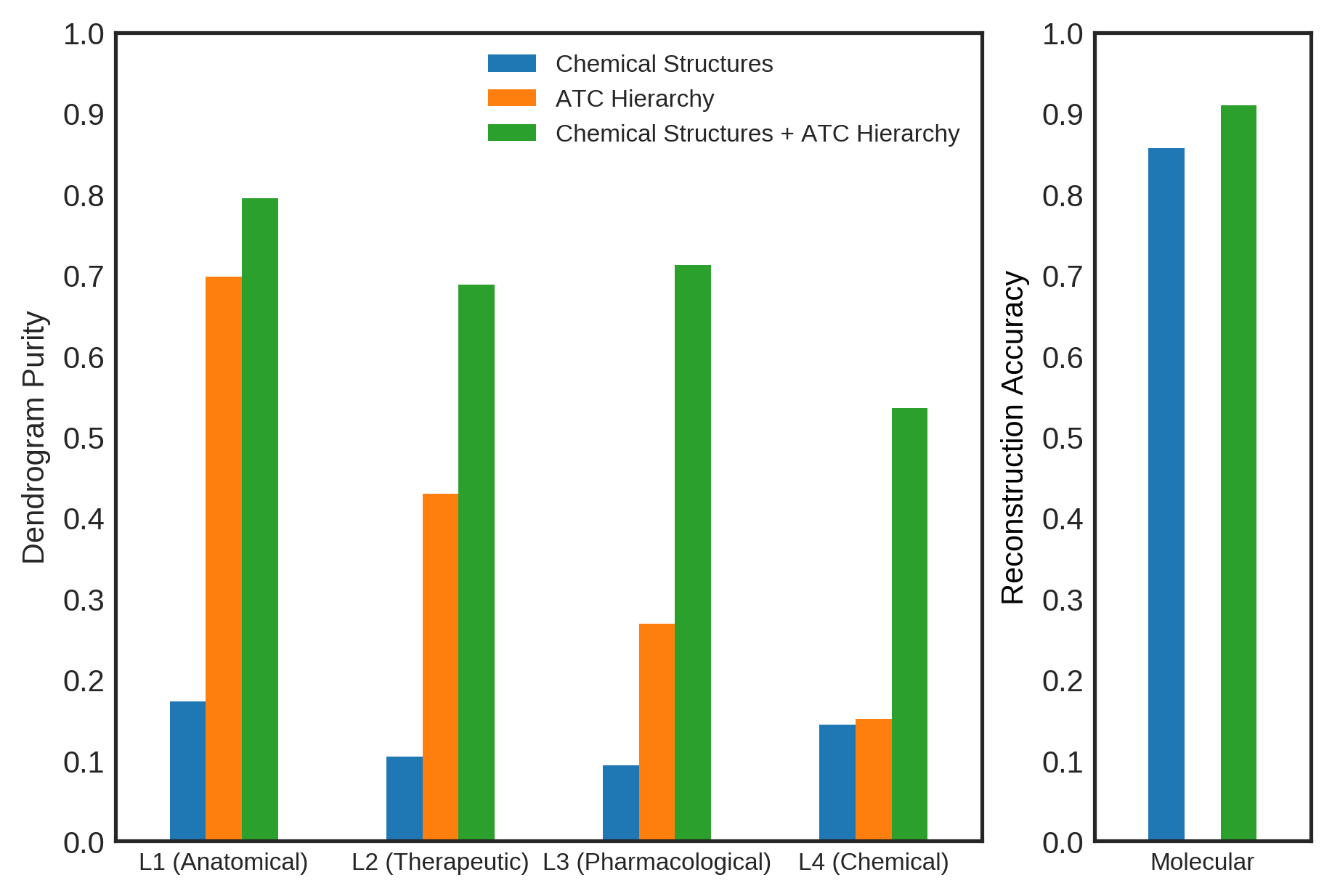}
    
    \caption{Effect of knowledge sources. Results obtained using the embedding from chemical structures alone are shown in blue, results obtained using the embedding from the ATC hierarchy alone are shown in orange, and results obtained using the embedding from both sources of knowledge are shown in green. The left panel shows the dendrogram purity (DP) at ATC levels 1, 2, 3 and 4. The right panel shows the reconstruction accuracy of the chemical structures. All results are from the Lorentz model with dimension size of 64.}
    
    \label{fig:task}
\end{figure}

The left panel in Figure~\ref{fig:task} shows DP at different ATC levels, and the right panel shows the reconstruction accuracy. The embedding obtained from both sources of knowledge has substantially better performance than embeddings derived from only one source of knowledge. The improvement in performance is particularly significant at the ATC levels 3 and 4 that cluster drugs by chemical structure. This result provides support that information learned from the task of molecular reconstruction can help inform the task of drug clustering and vice-versa.

\begin{table*}
\caption{Effect of hyperbolic space. Dendrogram purity (DP) values of drug hierarchies obtained using embeddings in hyperbolic space (Lorentz model) and in Euclidean space. DP values are shown at different ATC levels for both manifold geometries. The DP values are reported as the average of 3 independent runs with standard deviations. A boldface value indicates that the corresponding manifold geometry has a significantly higher DP value.}
\label{tbl:ecu_vs_lor}
\centering
\begin{tabular}{c c c c c c c}
\toprule
\multirow{2}{*}{\textbf{ATC Level}} & \multirow{2}{*}{\textbf{Geometry}} & \multicolumn{5}{c}{\textbf{Latent Space Dimension}} \\
\cmidrule{3-7}
& & 2 & 4 & 8 & 32 & 64 \\
\midrule
\multirow{2}{*}{L1 (Anatomical)} & Euclidean & $0.690_{\pm .013}$ & $0.721_{\pm .030}$ & $0.748_{\pm .029}$ & $0.774_{\pm .005}$ & $0.775_{\pm .008}$ \\ 
& Lorentz & $\bm{0.757_{\pm .006}}$ & $\bm{0.761_{\pm .014}}$ & $\bm{0.771_{\pm .006}}$ & $\bm{0.790_{\pm .003}}$ & $\bm{0.795_{\pm .001}}$\\

\midrule
\multirow{2}{*}{L2 (Therapeutic)} & Euclidean & $0.488_{\pm .023}$ & $0.626_{\pm .008}$ & $0.655_{\pm .017}$ & $0.681_{\pm .003}$ & $0.688_{\pm .003}$ \\ 
& Lorentz & $\bm{0.617_{\pm .007}}$ & $\bm{0.643_{\pm .015}}$ & ${0.666_{\pm .020}}$ & ${0.684_{\pm .007}}$ & ${0.690_{\pm .006}}$\\

\midrule
\multirow{2}{*}{L3 (Pharmacological)} & Euclidean & $0.384_{\pm .027}$ & $0.601_{\pm .018}$ & $0.668_{\pm .026}$ & $\bm{0.715_{\pm .006}}$ & $\bm{0.725_{\pm .006}}$ \\ 
& Lorentz & $\bm{0.577_{\pm .023}}$ & $\bm{0.641_{\pm .018}}$ & ${0.668_{\pm .018}}$ & ${0.696_{\pm .009}}$ & ${0.714_{\pm .001}}$\\

\midrule
\multirow{2}{*}{L4 (Chemical)} & Euclidean & $0.238_{\pm .022}$ & $0.402_{\pm .017}$ & $0.454_{\pm .017}$ & $\bm{0.597_{\pm .008}}$ & $\bm{0.625_{\pm .006}}$ \\ 
& Lorentz & $\bm{0.334_{\pm .007}}$ & $\bm{0.441_{\pm .010}}$ & ${0.457_{\pm .013}}$ & ${0.517_{\pm .006}}$ & ${0.528_{\pm .006}}$\\
\bottomrule
\end{tabular}

\centering
\end{table*}

\begin{figure*}
    \centering
    \includegraphics[width=1.0\textwidth]{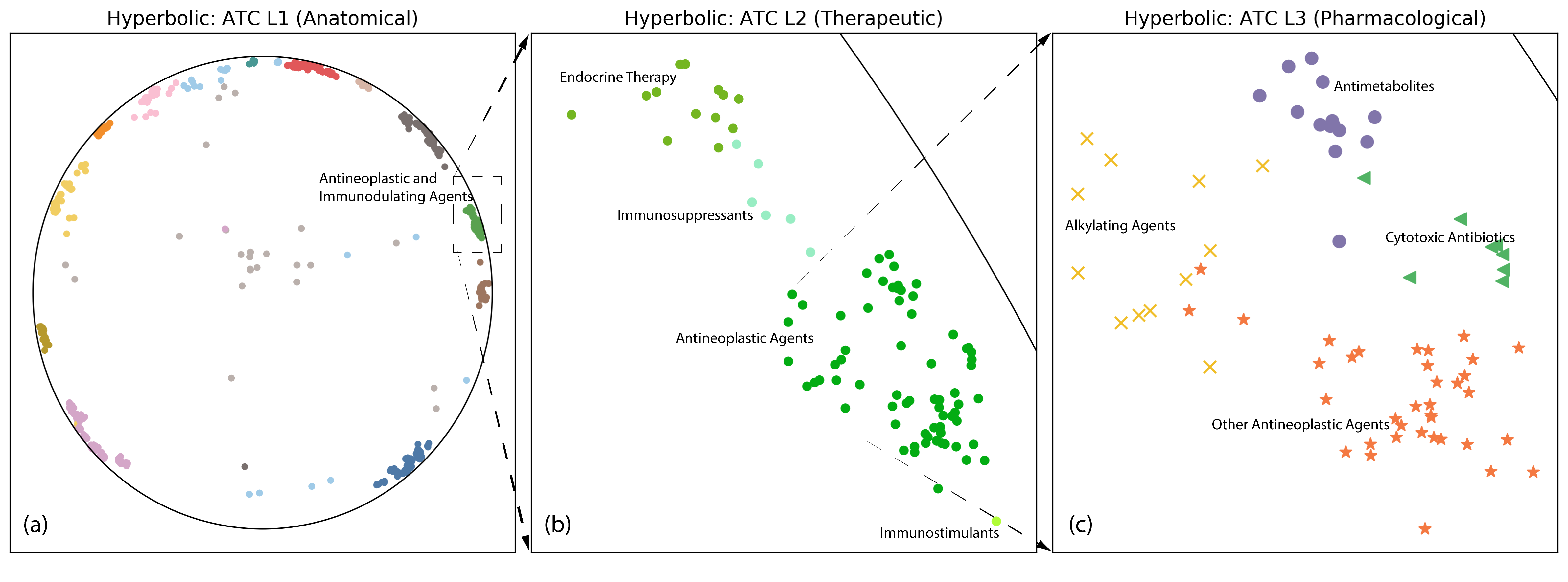}
    \caption{Visualization of hyperbolic drug embedding in two-dimensional Poincar\'e disk that shows drugs with colored symbols. In panel (a) drugs that belong to the same group at ATC level 1 are denoted by circles of the same color. Panel (b) shows drugs of one group from ATC level 1 namely, "Antineoplastic and Immunodulating Agents", and drugs that belong to the same group at ATC level 2 are denoted by circles with the same shade of green. Panel (c) shows drugs of one group from ATC level 2, namely, "Antineoplastic Agents", and drugs that belong to the same group at ATC level 3 are denoted by symbols of the same color.}
    
    \label{fig:emb_2d}
\end{figure*}

\textbf{Effect of hyperbolic space.}
We compare embeddings from the Lorentz model with embeddings from the Euclidean model. The results in Table~\ref{tbl:ecu_vs_lor} show that overall the Lorentz embeddings have higher DP values and outperform the Euclidean embeddings. In low dimensional spaces (dimension size of two to four), the Lorentz model produces higher-quality embeddings across all ATC levels, suggesting that hyperbolic space has superior capacity with the same dimension. In addition, the Lorentz model shows consistently higher DP values at ATC level 1, suggesting that it is superior to its Euclidean counterpart in recapitulating the global aspects of the hierarchy. Increasing the dimension of the latent space beyond 32 did not yield sizeable improvements at the ATC levels 1 and 2. For local aspects of hierarchy (ATC levels 3 and 4), the improvement from hyperbolic representation decreases as the latent dimensionality increases. Note that the results may be less reliable at ATC levels 3 and 4 due to the smaller sample sizes of the clusters. 

We visually explore the embedding in two dimensional hyperbolic space by mapping the embedding in the Lorentz model to the Poincar\'e disk via a diffeomorphism described in~\citep{nickel2018learning}. In Figure~\ref{fig:emb_2d}(a), we observe that most of the drugs are placed near the boundary of the Poincar\'e disk and form tight clusters that correspond to the drug groups at ATC level 1. The hyperbolic embedding exhibit a clear hierarchical structure where the clusters at the boundary can be viewed as distinct substrees with the root of the tree positioned at the origin. A small number of drugs (grey circles) are scattered around the origin and denote drugs that act on the on sensory organs or act on several organs. Figure~\ref{fig:emb_2d}(b) and (c) demonstrate that embedding in hyperbolic space can effectively induce a multi-level tree. More specifically, in Figure~\ref{fig:emb_2d}(b), we zoom into the level 1 group called ``Antineoplastic and Immunodulating Agents" and show that drugs form clusters that correspond to level 2 groups. We further zoom into one level 2 group called “Antineoplastic Agents” Figure~\ref{fig:emb_2d}(c) and demonstrate that members form clusters that correspond to level 3 groups. This example demonstrates that the embedding retains the hierarchical structure to the deepest levels.

\textbf{Summary.} The preceding results show that best performing embedding is obtained with the Lorentz model with dimension size of 64, when both the chemical structures and the ATC hierarchy are leveraged. 
We refer to this embedding as the Lorentz Drug Embedding (LDE) and we use it in the following experiments. 

\subsection{Predicting side effects}
Side effects are unwanted reactions to drugs and the occur commonly. Often, not all side effects of a drug are known at the time it is approved for medical use. Thus, it is of critical importance to identify side effects of drugs that are in use. We applied LDE to predict side effects and compared its performance to several state of the art drug representations for predicting side effects~\citep{wu2018moleculenet}. We apply the side effect prediction methods to predict the presence or absence of side effects of drugs in each of the 27 classes as defined in the SIDER database. We perform three independent runs with different random seeds. In each run, we randomly split the SIDER database into training, validation and test sets in the proportions 80\%:10\%:10\%.  We use mean AUC-ROC as the evaluation metric.

The comparison drug representations include (1) graph-based representations including Weave and Graph Convolutional (GC) network that represent each molecule as an undirected graph, (2) Fingerprint (FP) representation that is a fixed length binary encoding of topological characteristics of the molecule. For FP and LDE, we derive k-nearest neighbor (kNN) and random forest (RF) classifiers to predict side effects. We use Tanimoto coefficient~\citep{butina1999unsupervised} as the similarity metric for kNN+FP, the hyperbolic distance as the similarity metric for kNN+LDE, and choose $k=11$ for both kNN classifiers. We obtain the results for Weave, GC, and RF+FP from~\citep{wu2018moleculenet} since their experimental settings are the same as our settings.

\begin{figure}[h]
    \centering
    \includegraphics[width=0.8\textwidth]{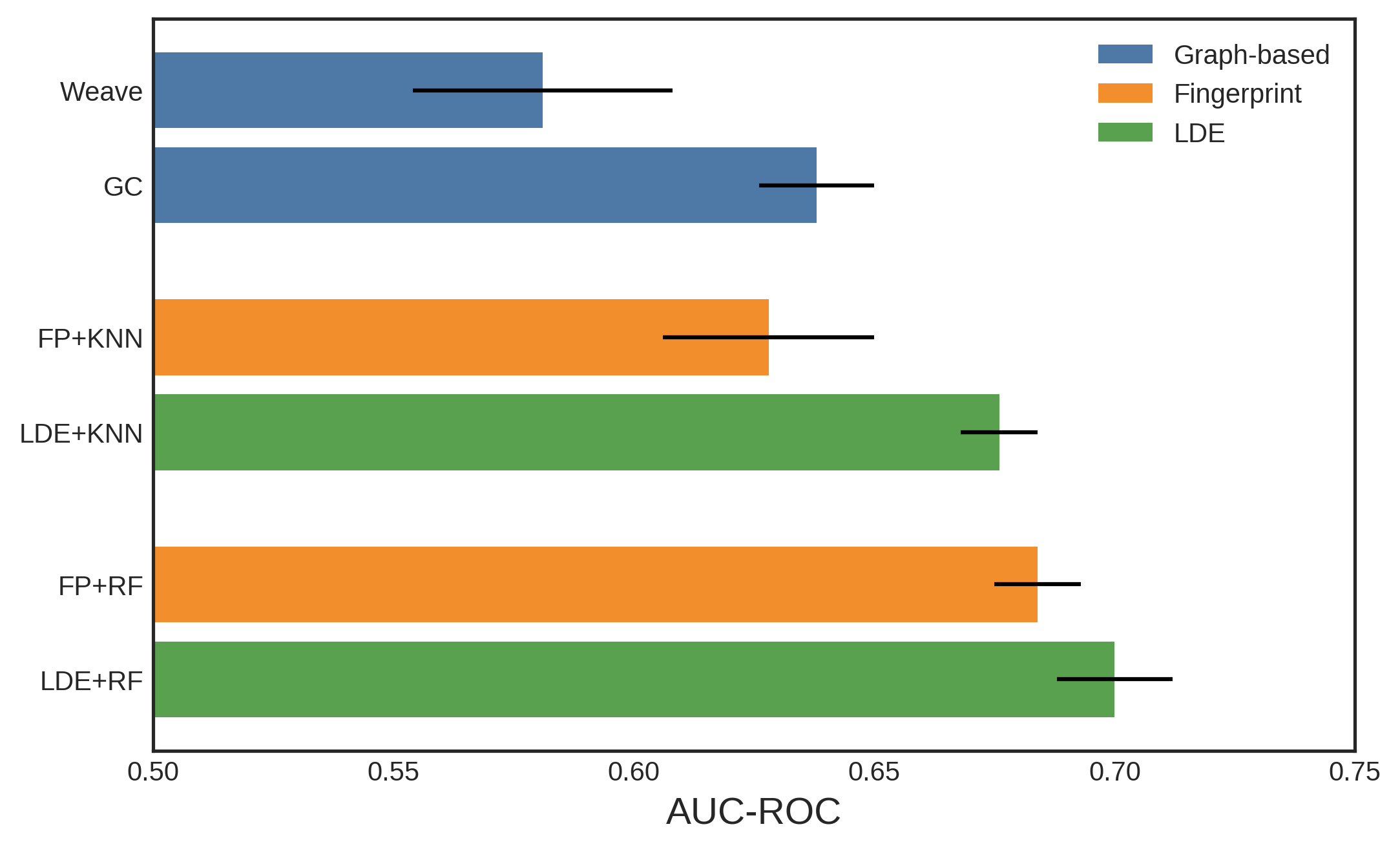}
    \caption{Prediction of side effects using three different representations of molecules: (1) Graph-based (Weave, GC), (2) Fingerprint (FP), and (3) Lorentz Drug Embeddings (LDE). The mean AUC-ROC values are the average of three independent runs and the error bars denote standard deviations.}
    \label{fig:SIDER}
\end{figure}

Figure~\ref{fig:SIDER} shows that LDE has significantly better performance in predicting side effects compared to both graph-based and FP representations.

\subsection{Evaluating drug repositioning}
Drug repositioning is the discovery of new uses, called indications, for approved drugs. Compared to \textit{de novo} drug discovery that takes enormous amount of time, money and effort, drug repositioning is more efficient since it takes advantage of drugs that are already approved. We evaluate LDE for drug repositioning by deriving kNN models to discriminate between approved and unapproved drug-indication pairs in the repoDB dataset. We tag each drug-indication pair with the date when the drug was first approved by the FDA. We choose 2000 as the cutoff year to split the repoDB data set into training (earlier than year 2000) and test (year 2000 and later) sets. For each drug $\bm{x}_i$ in the test set, we first encode it into the hyperbolic space using its SMILES string as the input, and then retrieve its $k$ nearest neighbors $\{X_{kNN}\}$ from the training set in the hyperbolic space. We apply majority voting to the retrieved drug-indication pairs in $\{X_{kNN}\}$ to predict the status of each indication associated with $\bm{x}_i$. For indications of $\bm{x}_i$ that do not exist in $\{X_{kNN}\}$, we assume that it has equal probability of being either being successfully approved or failed to be approved.

In Table~\ref{tbl:dr_cs}, we show an example of the drug esomeprazole as the query drug for which we want to predict new indications. Esomeprazole was first approved by FDA in 2001 and thus is not in the training set. The three most similar drugs to esomeprazole in the hyperbolic space are omeprazole, rabeprazole and famotidine, which were approved by FDA in 1989, 1999 and 1986 respectively. Table~\ref{tbl:dr_cs} shows that, based on the status of indications associated with the retrieved drugs, we successfully predict all uses of esomeprazole that have been approved by the FDA. Moreover, we observe that esomeprazole is not likely to be approved for nausea, laryngeal diseases, and cystic fibrosis based on the failed approval of omeprazole for these indications. 

\begin{table*}[t]
\caption{Example of drug repositioning prediction for esomeprazole using kNN (k=3). The first two columns show the ground-truth status of indications associated with esomeprazole. In the third column, a check mark represents one vote from a retrieved drug and a blank represents that the status of corresponding indication is absent in a retrieved drug.}
\label{tbl:dr_cs}
\centering
\begin{tabular}{c c c c c c}
\toprule
\multirow{2}{*}{\textbf{FDA Status}} & \textbf{Query Drug} & \multicolumn{3}{c}{\textbf{Retrieved Drugs}} \\
\cmidrule{3-5}
 & Esomeprazole & Omeprazole & Rabeprazole & Famotidine \\
\midrule
\multirow{5}{*}{Approved} & Erosive Esophagitis & \cmark & &\cmark \\
& Zollinger-Ellison Syndrome & \cmark & \cmark & \cmark\\
& Peptic Esophagitis & \cmark & & \cmark \\
& Gastroesophageal Reflux Disease & \cmark & \cmark & \cmark\\
& Peptic Ulcer	& \cmark & \cmark & \\
\midrule
\multirow{3}{*}{Unapproved} & Nausea & \cmark & & \\
& Laryngeal Diseases & \cmark & & \\
& Cystic Fibrosis & \cmark & & \\
\bottomrule
\end{tabular}
\centering
\end{table*}



\begin{figure}[h]
    \centering
    \includegraphics[width=0.8\textwidth]{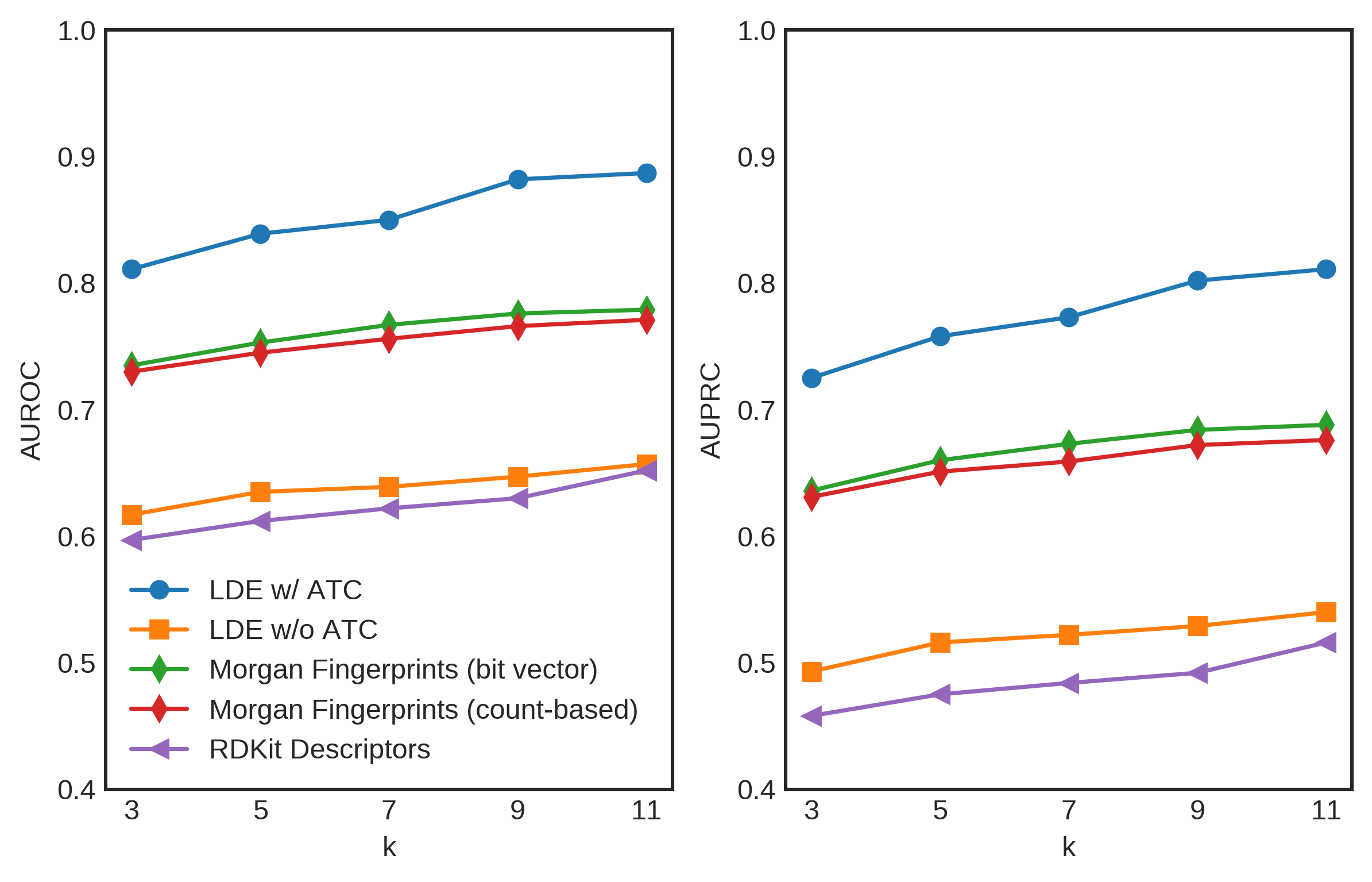}
    \caption{Comparison of representations for drug repositioning prediction using kNN (k $\in [3,5,7,9,11]$). The left panel shows AUROC scores and the right panel shows AUPRC scores.}
    \label{fig:repo_comp}
\end{figure}

Because we are not aware of any other approach developed on the repoDB dataset with the same chronological split, we compare the performance of LDE for drug repositioning using kNN, for each $k$ in $[3,5,7,9,11]$, to the following baselines: (1) kNN on RDKit-calculated descriptors, (2) kNN on Morgan (ECFP) fingerprints (bit vector of size 2048), (3) kNN on count-based Morgan fingerprints, and (4) kNN on Lorentz drug embedding without ATC information.
Performance is evaluated using area under the receiver operating characteristic curve (AUROC) and area under the precision-recall curve (AUPRC). Figure~\ref{fig:repo_comp} shows that the LDE with ATC information, i.e., pairwise similarity between drugs, outperforms other drug representations by a large margin. Averaging across different $k$ values, the LDE with ATC information surpasses Morgan (ECFP) fingerprints, the second best representation, by $12\%$ (AUROC) and $15.8\%$ (AUPRC). Compared to LDE without ATC information, incorporating drug hierarchy in the embedding achieves a large gain of $33.6\%$ (AUROC) and $48.8\%$ (AUPRC). LDE's competitive performance on discovering repositioning opportunities are likely driven by the drug-drug similarity that is encoded in the ATC hierarchy. 

\section{Conclusion}
We introduced a method for learning a high-quality drug embedding that integrates chemical structures of drug and drug-like molecules with local similarity of drugs implied by a drug hierarchy. We leveraged the properties of the Lorentz model of hyperbolic space and developed a novel hyperbolic VAE method that simultaneously encodes similarity from chemical structures and from hierarchical relationships. We showed empirically that our embedding possesses the hierarchical relationships in the ATC tree and can accurately reproduce the chemical structure. Our results support that learning chemical structure can help preserve the ATC hierarchy in the latent space and vice-versa. We further showed that the embedding can be used to discover new side-effects and for drug repositioning. 

There are several directions for future work. We hope to investigate the utility of integrating other sources of biomedical information, such as drug-target interaction data, into the model. Our approach is general and can easily incorporate additional knowledge based on the drug-drug similarity measurements in the dataset. Besides, as our framework is built on a probabilistic generative model, it would be interesting to investigate its utility for drug discovery, for example, searching new molecules that are similar to the FDA-approved drugs in a pharmacological class.

\bibliography{main}
\newpage
\section{Supplementary Material}
\noindent In this supplement, we provide an algorithmic description of our method and implementation details. We also provide additional ablation studies and two-dimensional visualizations of Euclidean drug embeddings. The datasets and Python scripts are available at the project repository https://github.com/batmanlab/drugEmbedding.

\subsection{Algorithmic description}
Algorithm~\ref{alg:main} provides the pseudocode for our method for learning drug embeddings using the Lorentz model (LDE).
\begin{algorithm*}[h]
\textbf{Input}: SMILES $X=\{X_\mathrm{ZINC}, X_\mathrm{FDA}\}$, drug classification system $\mathcal{T}$, sample per data point $L=1$, upsample weight $\omega$, scaling parameters $\beta$, $\gamma$, number of negative samples $K$, and positive example sampling strategy $\mathcal{S}$. \\
 \textbf{Initialize} $\phi$, $\theta$ \\
 \While{no convergence of $\phi, \theta$}{
    Sample minibatch $X^B=\{X^{B}_\mathrm{ZINC}, X^{B}_\mathrm{FDA}\}$ with $|X^B|=B$ and $|X^{B}_\mathrm{FDA}|:|X^B| = \omega$ \\
    Sample $\bm{u}^B \sim \mathcal{N} (0, \Sigma_\phi(X^B))$\\
    Reparameterization trick: $\bm{z}^B = g_\phi(\bm{u}^B, \bm{\mu}_\phi(X^B))$ by Eq.~(\ref{eq:hrep})\\
    Compute $\Tilde{\mathcal{L}}_{\beta-\mathrm{ELBO}}(X^B; \phi, \theta) = \frac{1}{B} \sum_i^B \big(\mathrm{log}p_\theta(\bm{x}_i|\bm{z}_i) + \beta \cdot  \Tilde{D}_\mathrm{KL}(q_\phi(\boldsymbol{z}|\boldsymbol{x}_i)\|p(\boldsymbol{z}))\big) $ by Eq.~(\ref{eq:KL_est})\\
    \For{ $i = 1, ..., B$}{
        \If{$\bm{x}_i \in X_\mathrm{FDA}$}{
        Sample $\mathcal{D}(i,j) \sim \mathcal{T}$, including one positive sample using strategy $\mathcal{S}$ and $K$ negative random samples\\
        Compute $\Tilde{\mathcal{L}}_{\mathrm{SLR}}(\bm{x}_i, \mathcal{D}(i,j); \phi)$ by Eq.~(\ref{eq:slr_est})
        }
    }
    Aggregate $\Tilde{\mathcal{L}}_{\mathrm{SLR}}(X_\mathrm{FDA}^B, \mathcal{T}; \phi) = \frac{1}{w \cdot B} \sum_i \Tilde{\mathcal{L}}_{\mathrm{SLR}}(\bm{x}_i, \mathcal{D}(i,j); \phi)$ \\
    Compute $\Tilde{\mathcal{L}}(X^B; \phi, \theta) = \Tilde{\mathcal{L}}_{\beta-\mathrm{ELBO}}(X^B; \phi, \theta) + \gamma \cdot \Tilde{\mathcal{L}}_{\mathrm{SLR}}(X_\mathrm{FDA}^B, \mathcal{T}; \phi) $\\
    Compute gradients $\nabla_{\phi, \theta}\Tilde{\mathcal{L}}(X^B; \phi, \theta)$ \\
    Update $\phi, \theta$ using stochastic gradient ascent
    }
 \caption{Learning drug embeddings in the Lorentz model (LDE)}
 \label{alg:main}
\end{algorithm*}

\subsection{Implementation details}
\label{sec:imp_details}
\subsubsection{Data preprocessing}
There may be many different SMILES strings that can be constructed for a given molecule by starting at a different atom or by following an alternative sequence through the molecule. We use the canonical SMILES strings that are generated by using the RDKit package~\cite{landrum2006rdkit}. Each $\bm{x} \in X$ is then formed as a sequence of one-hot vectors 
where each one-hot vector $x^{(j)}$ corresponds to the index of a symbol in the vocabulary. The vocabulary consists of all unique characters except for atoms represented by two characters, e.g., \texttt{Cl}, \texttt{Br}, and \texttt{Si} which are considered as single symbols. We limit the maximum length of SMILES and pad the shorter strings with special symbol \texttt{<pad>} so that the encoder network takes input sequences with a fixed length. 

\subsubsection{Model architecture}
We adopt the core structure of variational autoencoder used in the context of written English sentences~\cite{bowman2015generating}. More specifically, we use single-layer GRU RNNs~\cite{cho2014learning} with 512 hidden units for both encoder $q_\phi(\bm{z}|\bm{x})$ and decoder $p_\theta(\bm{x}|\bm{z})$ networks. To assure the output $\bm{\mu}$ of the encoder is on the hyperbolic manifold $\mathbb{H}^n$, we apply $\mathrm{exp}_{\bm{\mu}_0}$ to the final layer of the encoder. The decoder serves as a language model for SMILES to 120 and is trained with teacher forcing~\cite{williams1989learning}. We use the Adam optimizer~\cite{kingma2014adam} with $\beta_1 = 0.9$, $\beta_2=0.999$, and use a learning rate of 0.003. We implement our models using PyTorch~\cite{paszke2017automatic} version 1.0.0.

\subsubsection{Training settings}
We use 128 samples per mini-batch for all experiments. Since the data set $X=\{X_\mathrm{ZINC}, X_\mathrm{FDA}\}$ is highly unbalanced with $|X_\mathrm{ZINC}| \gg |X_\mathrm{FDA}|$, we use a weighted random sampler to add more examples from the FDA-approved drugs during training. Based on our empirical results, we find that upsample $\bm{x} \in X_\mathrm{FDA}$ to $20\%$ in each mini-batch is sufficient for learning the drug hierarchy.

We notice that, in the standard setting, the VAE suffers from the ``posterior collapse'' issue~\cite{van2017neural}, where the model learns to ignore the latent variable and approximate posterior mimics the prior. To mitigate this issue, we employ a KL-annealing schedule, by which a variable $w_t$, increasing from zero to one in a ``warm-up" period, is applied to the KL divergence term. In addition, following~\cite{kusner2017grammar}, we set $\beta$ in the loss function to $1/dim(\bm{z})$, which lessens the effect of KL divergence term and encourages learning latent representations with high reconstruction accuracy.

Meanwhile, we fine-tune two hyperparameters: (1) $\gamma$, the weight of the soft local ranking loss, and (2) $K$, number of randomly sampled negative examples for each anchor data point. Both of them affect the model's capability of learning hierarchy. We perform ablation studies of these two hyperparameters and find that using $\gamma = 11$ and $K=11$ are good options in our experiments.

\newpage
\subsection{Weight of the soft local ranking and the number of negative samples}
\begin{figure}[h]
    \centering
    \includegraphics[width=0.8\textwidth]{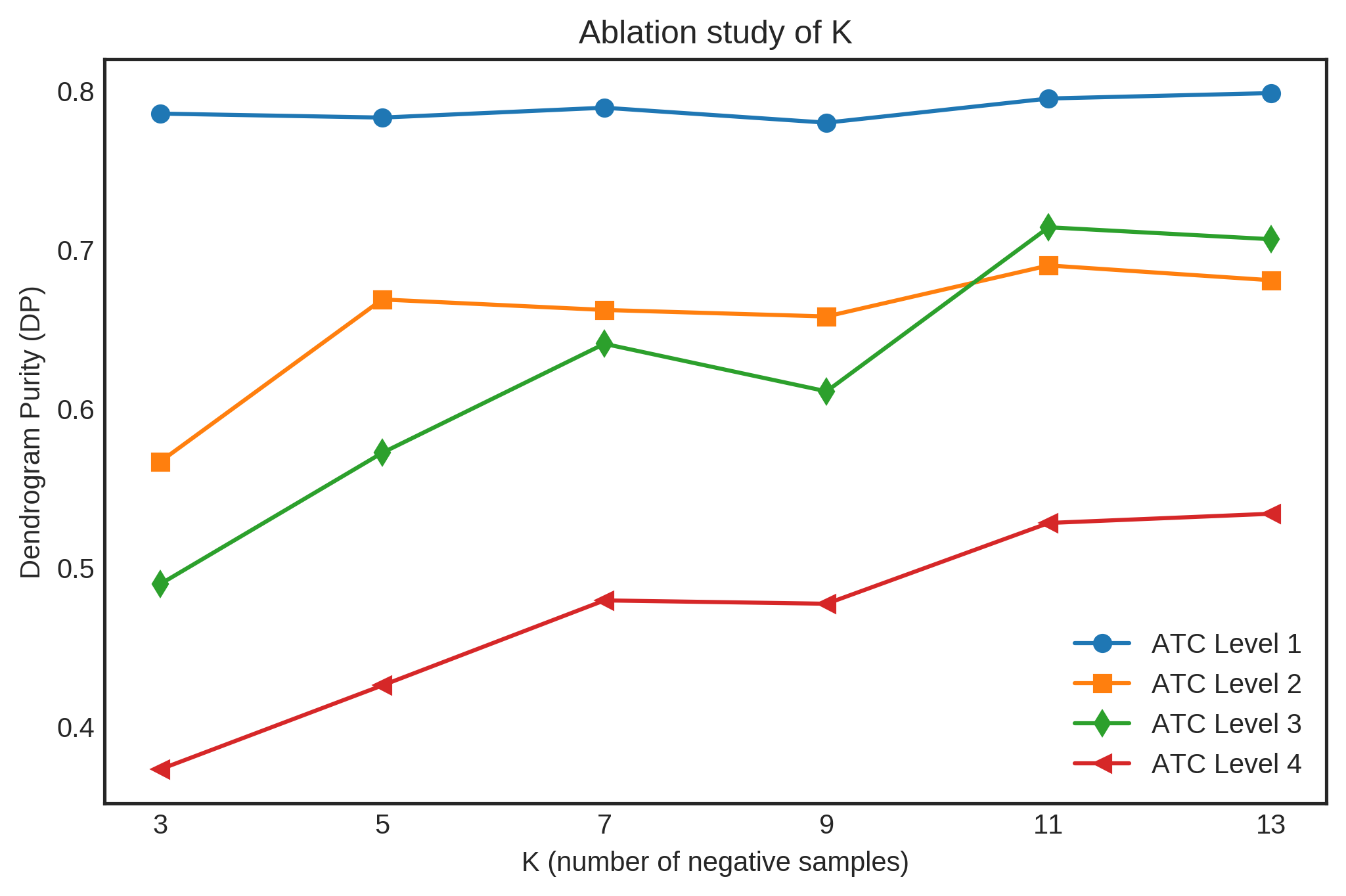}
    \caption{Ablation study on sampling size of negative examples ($K$). Dendrogram purities of the ATC level 1, 2, 3, 4 at epoch 100 are plotted against $K$. We set $\gamma=K$ in these experiments. All results are from the Lorentz model with dimension size of 64.}
    \label{fig:abl_k}
\end{figure}

In the Figure~\ref{fig:abl_k}, we demonstrate that using larger $\gamma$ and larger $K$ can accelerate the learning of the drug hierarchy. However, increasing $\gamma$ and $K$ beyond 11 does not yield further improvements across all the ATC levels. Based on the results, we chose $\gamma=11, K=11$ in our study.   

\newpage
\subsection{Effect of sampling strategies}
\begin{figure}[h]
    \centering
    \includegraphics[width=0.8\textwidth]{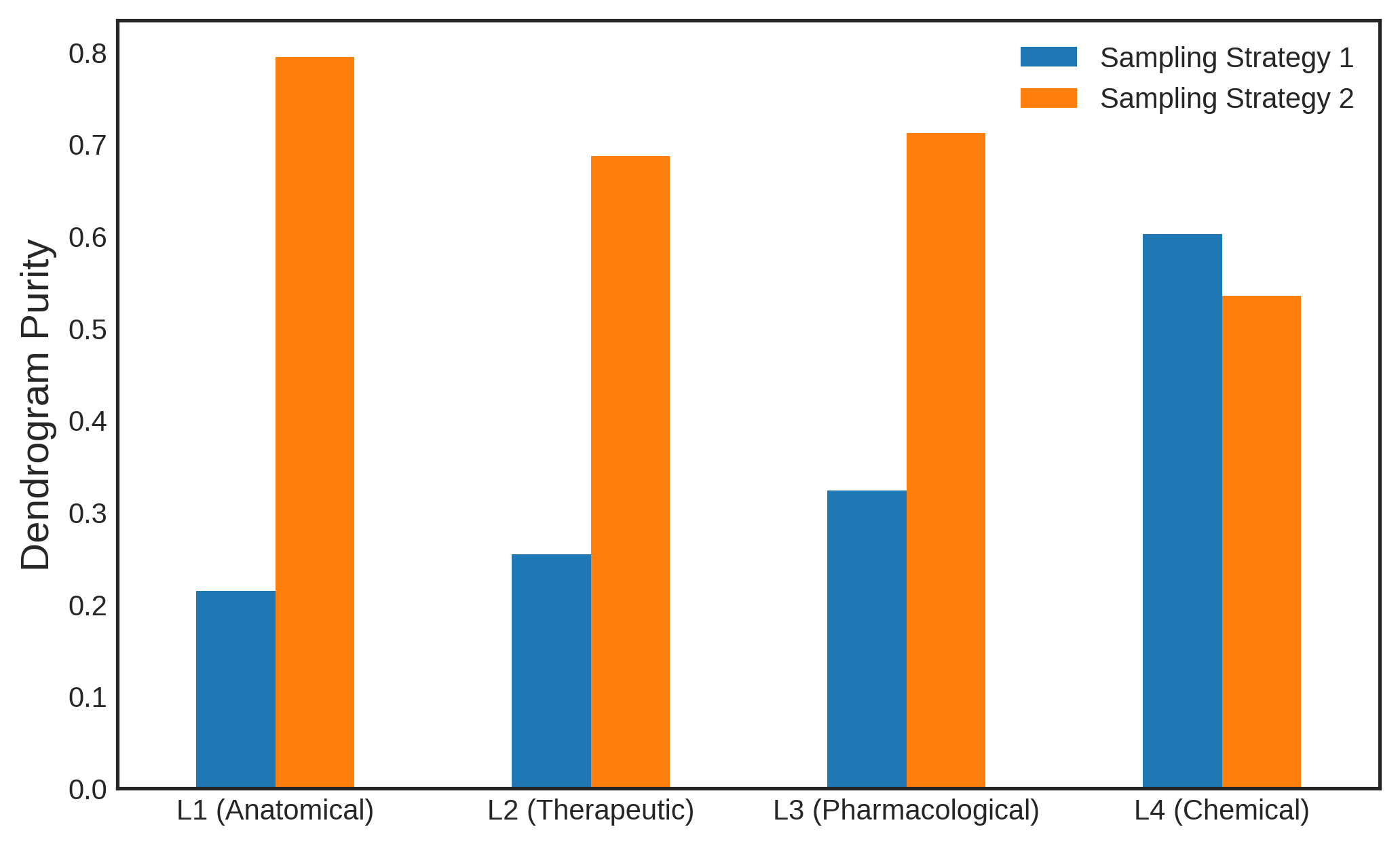}
    \caption{Effect of sampling strategies. Dendrogram purity (DP) values obtained using two sampling strategies for learning drug hierarchies. All results are from the Lorentz model with dimension size of 64.}
    \label{fig:ss}
\end{figure}

As stated in Section~\ref{sec:slr}, effective sampling of $\mathcal{D}(i,j) \sim \mathcal{T}$ is critical for learning the underlying drug hierarchy. In Figure~\ref{fig:ss}, we demonstrate two sampling strategies that result in very different performances. Recall that $\mathcal{D}(i,j)=\{k: t_{i,j}<t_{i,k}\} \cup \{j\}$ is the set of drugs that have path-lengths longer than or equal to $t_{i,j}$. In the first sampling strategy, for each anchor example $\bm{x}_i$, we sample the positive example $\bm{x}_j$ from its nearest neighbors, while in the second sampling strategy, we uniformly sample a positive example $\bm{x}_j$, such that $\bm{x}_j$ has an equal chance of being drawn from any subtree including $\bm{x}_i$. In both sampling strategies, negative examples $\bm{x}_k$ are randomly sampled from more distant terminal nodes, i.e., $t_{i,k}>t_{i,j}$. Intuitively, the first sampling strategy focuses attention on the local aspects of the hierarchy, while the second strategy emphasizes the global aspects of the hierarchy by setting up more difficult classification tasks. Figure~\ref{fig:ss} shows that the second sampling strategy yields substantially better DP values at higher ATC levels at the cost of poorer performance at ATC level 4. 

\newpage
\subsection{Euclidean 2D drug embeddings}
\begin{figure*}[h]
    \centering
    \includegraphics[width=0.9\textwidth]{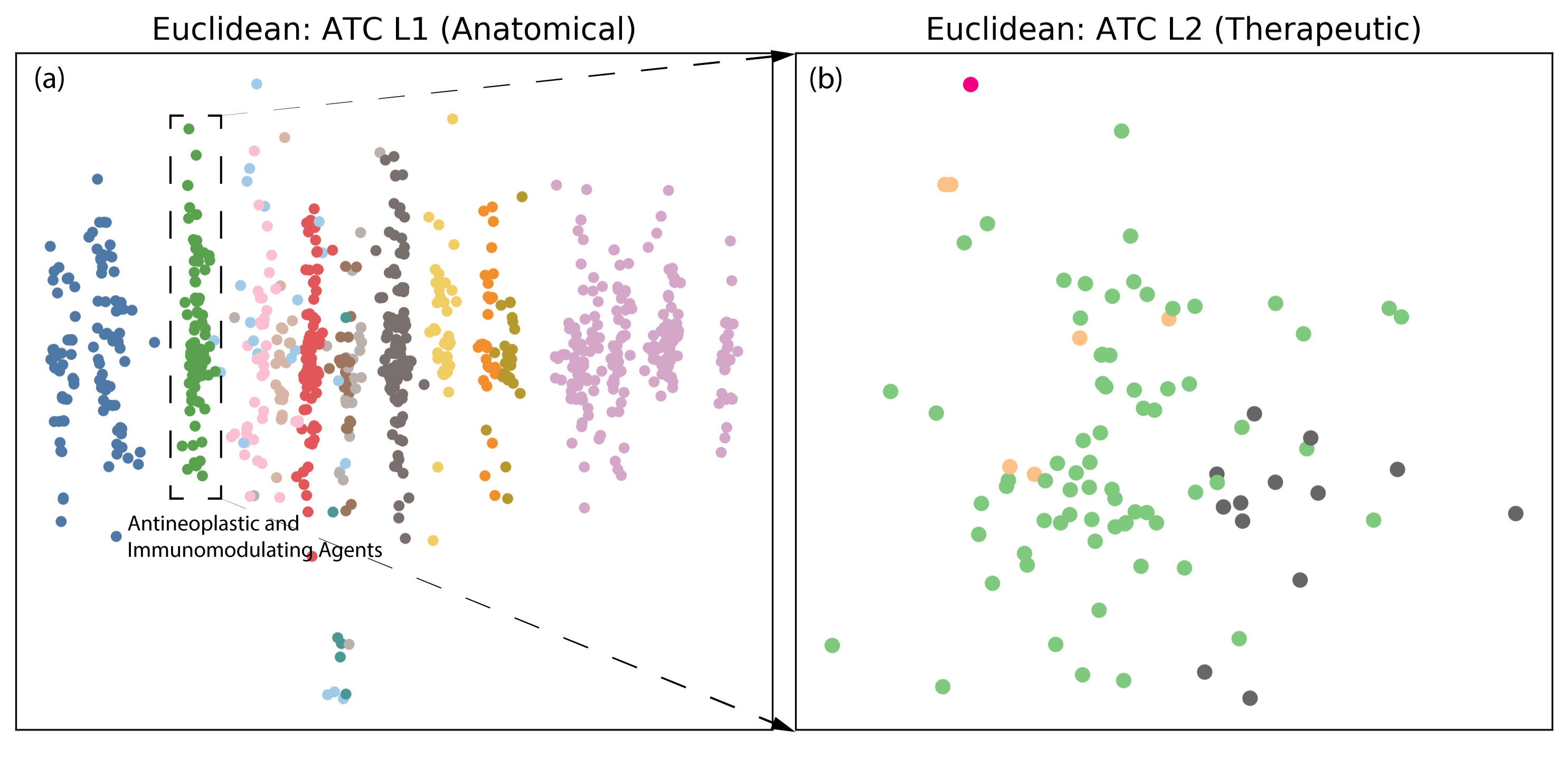}
    \caption{Visualization of drug embeddings in two-dimensional Euclidean space. Panel (a) shows drugs at ATC level 1 and drugs in same group are denoted by circles of the same color. Panel (b) shows the details of one ATC level 1 group, namely ``Antineoplastic and Immunomodulating Agents", and drugs in the same subgroup at ATC level 2 are denoted by circles with the same color.}
    \label{fig:euc_2d}
\end{figure*}

[h]
Figures~\ref{fig:euc_2d} shows the Euclidean counterparts of Figures~\ref{fig:emb_2d}. In Figure~\ref{fig:euc_2d}(a), we observe that clusters in Euclidean space are parallelly partitioned and do not exhibit a multi-level tree structure. In Figure~\ref{fig:euc_2d}(b), we observe that embeddings of drugs in different therapeutic subgroups of ``Antineoplastic and Immunomodulating Agents" start to mix together, showing worse clustering performance than the embeddings in the hyperbolic space.

\newpage
\subsection{Drug repositioning case study}
\begin{figure}[h]
    \centering
    \includegraphics[width=0.7\textwidth]{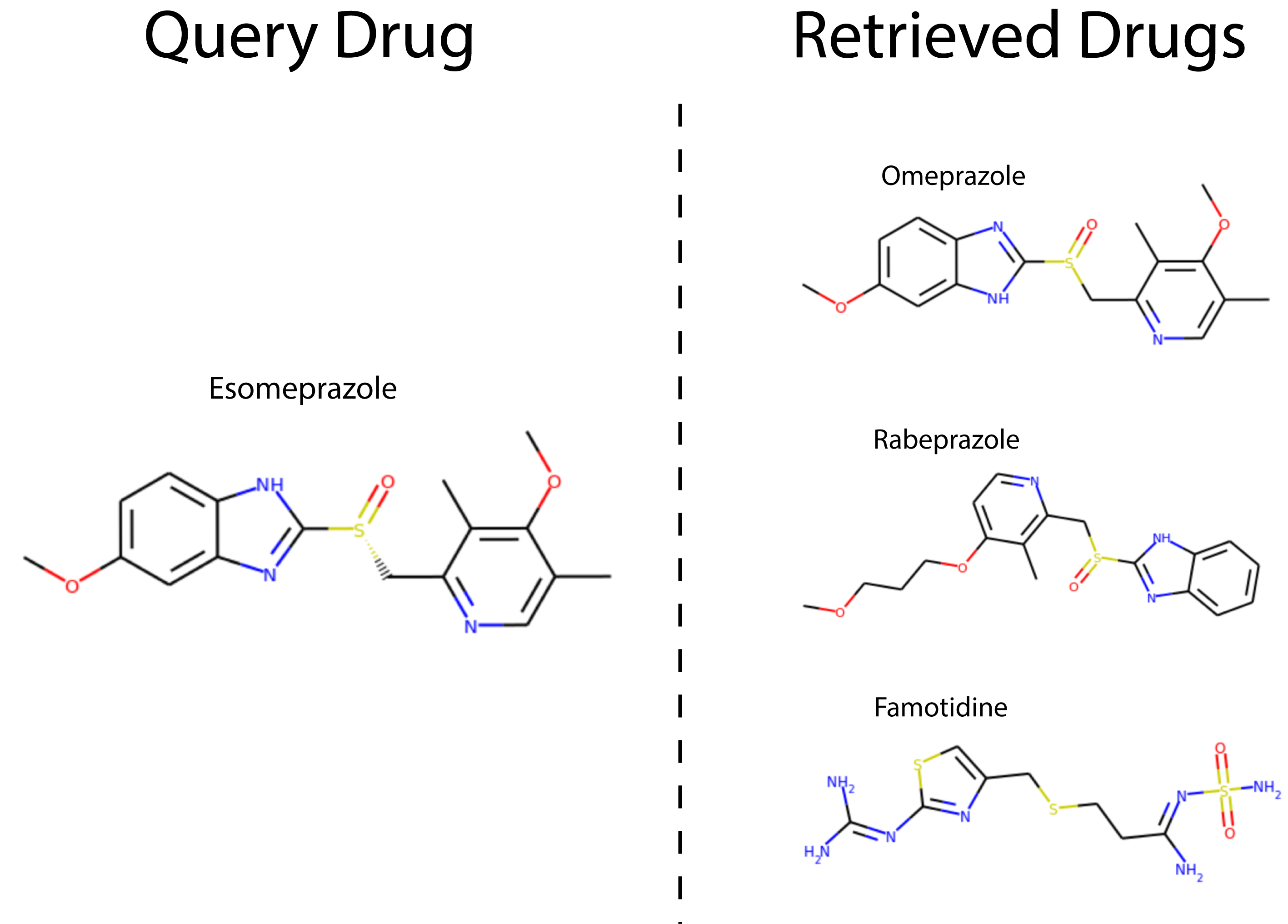}
    \caption{Molecular structures of esomeprazole and its 3 nearest neighbors retrieved using kNN. Among the retrieved drugs, omeprazole is closely related to esomprazole in chemical structure and rabeprazole shares a sub-structure with esomprazole. Although famotidine is structurally different, it belongs to the same pharmacological group as omeprazole and rabeprazole in the ATC hierarchy.}
    \label{fig:dr_cs}
\end{figure}

Figure~\ref{fig:dr_cs} shows the two-dimensional molecular structures of esomeprzole and its 3 nearest neighbors.

\end{document}